\documentclass[10pt,journal,compsoc]{IEEEtran}

\usepackage{mathtools}
\usepackage{xcolor}
\usepackage{graphicx}
\usepackage{amsmath}
\usepackage{indentfirst}
\usepackage{array, multirow, multicol}
\usepackage{caption}
\usepackage{tabularx,ragged2e}
\usepackage{algorithm}
\usepackage{algorithm}
\usepackage{algpseudocode}
\usepackage{amsmath}
\usepackage{graphics}
\usepackage{epsfig}
\usepackage{stackengine}

\algnewcommand\algorithmicoutput{\textbf{Until} convergence}

\usepackage{algpseudocode}
\usepackage{booktabs}

\ifCLASSOPTIONcompsoc
  \usepackage[nocompress]{cite}
\else
  \usepackage{cite}
\fi

\algnewcommand\Output{\item[\algorithmicoutput]}

\ifCLASSINFOpdf

\else

\fi

\usepackage{array}

\begin{document}
\title{Adversarial Information Bottleneck}
\author{Penglong~Zhai~\IEEEmembership{}and
        Shihua~Zhang~\IEEEmembership{}
\IEEEcompsocitemizethanks{\IEEEcompsocthanksitem Penglong Zhai and Shihua Zhang are with Academy of Mathematics and Systems Science, Chinese
Academy of Sciences, Beijing 100190, China, and School of Mathematical
Sciences, University of Chinese Academy of Sciences, Beijing 100049,
China.
\protect\\
Correspondence should be addressed to S.Z. Email: zsh@amss.ac.cn}

}



\IEEEtitleabstractindextext{
\begin{abstract}
The information bottleneck (IB) principle has been adopted to explain deep learning in terms of information compression and prediction, which are balanced by a trade-off hyperparameter. How to optimize the IB principle for better robustness and figure out the effects of compression through the trade-off hyperparameter are two challenging problems. Previous methods attempted to optimize the IB principle by introducing random noise into learning the representation and achieved state-of-the-art performance in the nuisance information compression and semantic information extraction. However, their performance on resisting adversarial perturbations is far less impressive.
To this end, we propose an adversarial information bottleneck (AIB) method without any explicit assumptions about the underlying distribution of the representations, which can be optimized effectively by solving a Min-Max optimization problem. Numerical experiments on synthetic and real-world datasets demonstrate its effectiveness on learning more invariant representations and mitigating adversarial perturbations compared to several competing IB methods.
In addition, we analyse the adversarial robustness of diverse IB methods contrasting with their IB curves, and reveal that IB models with the hyperparameter $\beta$ corresponding to the knee point in the IB curve achieve the best trade-off between compression and prediction,  and has best robustness against various attacks.
\end{abstract}

}

\maketitle

\IEEEdisplaynontitleabstractindextext

%
\IEEEpeerreviewmaketitle

\IEEEraisesectionheading{\section{Introduction}\label{sec:introduction}}

\IEEEPARstart{D}{eep} learning has achieved impressive successes in computer vision \cite{He2016DeepRL, Huang2017DenselyCC}, speech recognition \cite{AbdelHamid2014ConvolutionalNN}, competitive game playing \cite{Silver2017MasteringCA}, bioinformatics \cite{Cao2019SimpleTO} and so on. However, it turns out that deep neural networks (DNNs) are actually quite brittle, and particularly vulnerable to adversarial examples. Tiny noises in the input pixels is accumulated layer by layer in DNNs, which destructively destroy the carefully designed and well trained models. How to learn representations that are insensitive (invariant) to nuisances such as translations, rotations, occlusions is still a challenging task.
From the information-theoretic view, overfitting occurs when nuisance or irrelevant information are memorized by reducing empirical loss, rather than learning the true general relationship between the data $X$ and the labels $Y$ \cite{Krueger2017DeepND,Bashir2020AnIP}. Therefore, better generalization and robustness depend on whether the model memorizes more semantic or relevant information and compresses as much nuisance information as possible. This information-theoretic insight on deep learning can be formulated as the trade-off between information compression and prediction.
Besides, it is often concerned that DNNs are uninterpretable or not human-explainable \cite{10.1093/nsr/nwaa159}. For example, the lack of interpretability is particularly problematic in medical contexts, where safety risks can arise when there is mismatch between how a model is trained and used. The forms the explanations often taken are what the representations learn from the input data and how the network will respond to small perturbations \cite{Ribeiro2016WhySI,ShwartzZiv2017OpeningTB}.

\begin{figure}
  \centering
  \includegraphics[width=0.48\textwidth]{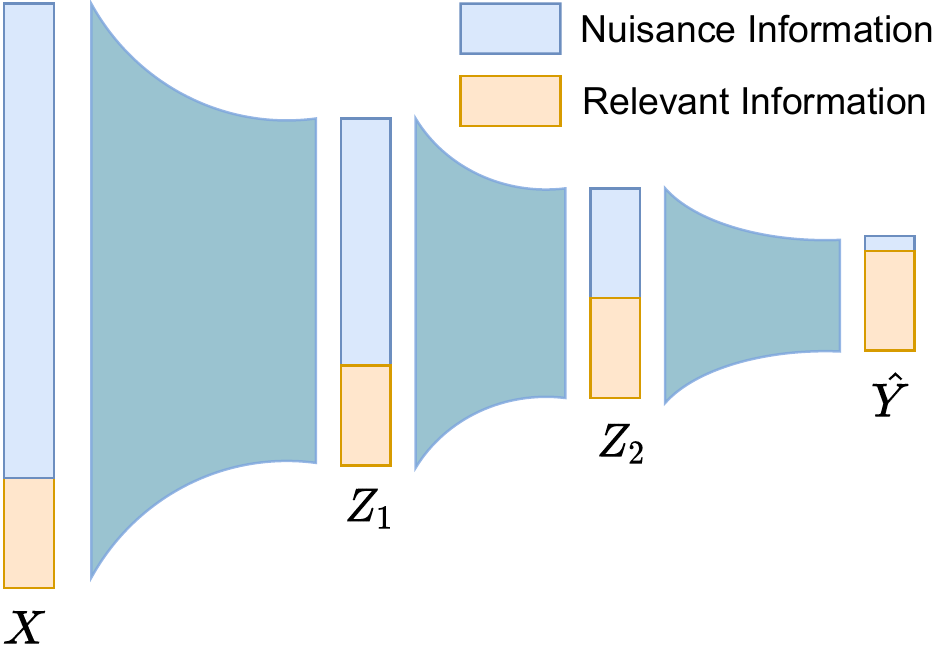}
  \caption{Schematic diagram of successive extraction of relevant information in DNN. The network receives an input $X$, and successively processes it through the layers $Z_{i,1 \leq i \leq L}$ (here $L=2$) to the predicted output $\hat{Y}$. Nuisance information denotes the mutual information between the layer representation and nuisance information, and relevant information denotes the mutual information between the layer representation and relevant information.}
\label{img:fig1}
\end{figure}

In general, DNN receives an input $X$, and successively processes it through the layer output $Z_{i}$ ($1 \leq i \leq L$, where $L$ is the number of hidden layers) to the predicted output $\hat{Y}$, which form a Markov chain and obeys the data processing inequality (DPI),
\begin{equation}
\label{eq:DPI}
\mathrm{H}(X) \geq \mathrm{MI}(X; Y) \geq \mathrm{MI}(Z_i; Y) \geq \mathrm{MI}(\hat{Y};Y)
\end{equation}
where $\mathrm{MI(\cdot; \cdot)}$ denotes the mutual information between two random variables. DPI guarantees that any information the hidden layer $Z_i$ has about $Y$ is extracted from $X$, and the deeper layers have less information for prediction (Fig. \ref{img:fig1}). However, with layer getting deeper, relevant information is expected to be retained while nuisance information is expected to be compressed. The fraction $\frac{\mathrm{MI}(Z; Y)}{\mathrm{MI}(X; Y)}$ quantifies how much of the relevant information is captured by DNN, and is expected to be maximized. This is precisely the target of the information bottleneck (IB) principle \cite{Tishby2015DeepLA}.

The IB principle aims to find a representation bottleneck variable $Z$, by maximizing the prediction, formulated in terms of $\mathrm{MI}(Z; Y)$, given a constraint on the compression, formulated in terms of $\mathrm{MI}(X; Z)$. Formally, this can be formulated into the following constrained optimization problem,
\begin{equation}
\begin{array}{ll}
\label{eq:constrained_IB}
&\max_{Z} \mathrm{MI}(Z; Y) \\
&\textsc{s.t.} \quad \mathrm{MI}(X; Z)\leq C
\end{array}
\end{equation}
where $C$ is the predefined compression rate, i.e., the minimal number of bits needed to describe the data. In practice, optimal bottleneck variables are usually not found by solving the above constrained optimization problem, but rather by maximizing the IB Lagrangian \cite{GiladBachrach2003AnIT, Shamir2010LearningAG},
\begin{equation}
\mathcal{L}_{IB}(Z) = \mathrm{MI}(Z; Y) - \beta \mathrm{MI}(X; Z)
\label{eq:IB_lagrangian}
\end{equation}
$\mathcal{L}_{IB}(Z)$ is the Lagrangian relaxation of the constrained optimization problem, and $\beta$ is a Lagrange multiplier that enforces the constraint $\mathrm{MI}(X; Z) \leq C$, which can also be taken as the hyperparameter that balances the fitting ability (prediction) and representation invariance (compression) in supervised learning setting.
Following recent information-theoretic techniques from \cite{He2016DeepRL}, Bassily \emph{et al.} \cite{Bassily2018LearnersTU} proved the following inequality
\begin{equation}
P\left[\left|\mathrm{err}_{\text {test }}-\mathrm{err}_{\text {train }}\right|>\epsilon\right]<O\left(\frac{\mathrm{MI}(X; Z)}{n \epsilon^{2}}\right)
\end{equation}
where $\mathrm{err}_{\text {test }}$ and $\mathrm{err}_{\text {train }}$ are the test error and the training error, respectively, $n$ is the training size, and $\epsilon \textgreater 0$ is a positive real number.
The intuition behind this inequality is that, the more a learning algorithm uses bits of the training set, there is potentially more overfitting risk. Thus, minimizing $\mathrm{MI}(X; Z)$ will have potential to avoid overfitting. Clearly, optimizing the IB Lagrangian not only minimize it, but also maximize prediction ability.
However, the fitting ability and representation invariance cannot be optimized simultaneously for a specific DNN model. This can be visualized on the IB curve \cite{Tishby2000TheIB,phdthesis}, which characterizes the set of bottleneck variables that achieve maximal $\mathrm{MI}(Z; Y)$ for a given $\mathrm{MI}(X; Z)$ by maximizing the IB Lagrangian given various $\beta$. In particular, visualizing the DNNs in terms of information theory is capable of helping researchers to detect erroneous reasoning in classification problems.

The IB methods have been found useful in a wide variety of learning applications (e.g., word clustering \cite{Slonim2000DocumentCU}, image clustering \cite{Goldberger2002UnsupervisedIC}). In particular, the IB principle has been employed to interpret deep learning as a successive information extraction process \cite{Tishby2015DeepLA}. Specifically, IB has been used to analyze DNNs by computing mutual information between the hidden layers and data or labels. This has attracted tremendous attention recently as a tool to gain insights into the learning dynamics and generalization ability of DNNs \cite{ShwartzZiv2017OpeningTB,pmlr-v97-goldfeld19a,Saxe2018OnTI}.
Besides, recent studies have demonstrated the effectiveness on improving generalization ability by optimizing the IB Lagrange empirically  \cite{Alemi2017DeepVI},\cite{Kingma2014AutoEncodingVB},\cite{chalk2016relevant},\cite{Achille2018InformationDL}, \cite{Amjad2020LearningRF}. For example, Alemi \emph{et al.} \cite{Alemi2017DeepVI} presented a variational approximation to IB (VIB), which is an adaption of Variational Auto-Encoder (VAE) \cite{Kingma2014AutoEncodingVB} to a prediction task. Numerical experiments showed that VIB had better generalization performance compared with several competing regularization methods, like dropout. Subsequently, Matthew \emph{et al.} \cite{chalk2016relevant} improved VIB by introducing different surrogate marginal distributions (e.g., Student$^{'}$s t-distribution) on the intermediate representation layer. Achille and Soatto \cite{Achille2018InformationDL} approximated the IB Lagrangian by introducing a information dropout layer, which allows the network to selectively introduce multiplicative noise into the layer activation, and thus to control the flow of information. Kolchinsky \cite{Kolchinsky2019NonlinearIB} introduced a novel non-parametric upper bound for mutual information by injecting noise to the representation layer, which achieved the best trade-off between compression and prediction. These existing methods optimized the IB Lagrangian by introducing random noise into learning the  representation, and achieved state-of-the-art performance in terms of nuisance information compression and relevant information extraction. However, their performance on resisting adversarial perturbations is far less impressive. We attribute this counter-intuitive result to the noise introduced in the IB methods. In addition, how to trade off the information compression and prediction for better adversarial robustness is a challenging issue and has not been explored so far.

We should note that the IB principle is not the only tool to encourage the representation invariance (proper compression) and adversarial robustness. For example, Achille and Soatto \cite{Achille2018EmergenceOI} showed that invariance to nuisance is equivalent to the information minimality of the learned representation and proposed to bound the model complexity using the information in the weights. More recently, Yu \emph{et al.} proposed an information-theoretic measure that maximizes the coding rate difference between the whole data and the sum of each individual class \cite{Yu2020LearningDA}.
Besides, bulk of non-information-theoretical principles have been proposed for reducing the sensitivity of DNN to small perturbations. Among them, double backpropagation is an old idea originally introduced by Drucker and Le Cun \cite{165600}, which trains the DNNs by minimizing not just the ``energy" of the network but the rate of change of that energy with respect to the input data. Xu and Mannor \cite{Xu2011RobustnessAG} proved that the Lipschitz constant of DNNs controls the difference between the training loss and generalization performance, and proposed to minimize it to decrease the model sensitivity to adversarial examples. In addition, Rifai \emph{et al.} \cite{Rifai2011ContractiveAE} proposed to penalize the representation sensitivity, measured as the squared norm of its Jacobian with respect to the input data. The Jacobian constraint has been utilized to learn contractive representations in supervised and unsupervised tasks \cite{Rifai2011ContractiveAE,Gu2015TowardsDN}. However, these hard-constraint methods lack of interpretability and meaningful measure like mutual information in IB to demonstrate their learning effectiveness.

In this paper, we propose adversarial information bottleneck (AIB) to optimize the IB Lagrangian by introducing an adversarial regularization term to approximate the information compression term $\mathrm{MI}(X; Z)$. AIB does not impose any explicit assumptions to the distribution of the hidden representation and can be optimized effectively by solving a Min-Max optimization problem. Numerical results demonstrate its effectiveness on the nuisance information compression and adversarial robustness.
In addition, we visualize the IB curves of several competing IB methods by varying $\beta$ on synthetic and real-world datasets, and reveal that all the IB curves present two distinct phases. By analyzing the adversarial robustness contrasting with the IB curves, we empirically claim that the models trained with the trade-off hyperparameter corresponding to the knee point in the IB curves present the best robustness.

\section{Related Work}\label{sec:related}
\subsection{Information Bottleneck Principle}
Tishby \emph{et al.} \cite{tishby2000information} firstly proposed the IB concept and provided a tabular method based on the Blahut-Arimoto (BA) algorithm \cite{Tishby2000TheIB, 1405276} to numerically solve the IB Lagrangian (Eq. \ref{eq:IB_lagrangian}) for the optimal encoder distribution $P(Z|X)$, given the trade-off hyperparameter $\beta$ and the cardinality of the representation. However, the BA algorithm can only be used in two special cases: the first is where $X$ and $Y$ are discrete-valued with a small number of possible states, and the second is when $X$ and $Y$ are continuous-valued and jointly Gaussian \cite{Chechik2003InformationBF}. In these two cases, the conditional probability can be computed explicitly during the optimization process. However, this is infeasible in the big data era due to the high-dimensionality of data.

More recently, Tishby \emph{et al.} interpreted deep learning as a successive information-extraction process quantified by the IB principle \cite{Tishby2015DeepLA} and attempted to open the black box of deep learning by visualizing the behavior of the compression and prediction terms \cite{ShwartzZiv2017OpeningTB}. This has motivated many studies to apply the IB principle to high-dimensional and complex data in a variety of scenarios such as improving the robustness against adversarial attacks \cite{Fischer2020TheCE}, learning invariant and disentangled representations \cite{Achille2018EmergenceOI} and interpreting the learning dynamics of DNNs \cite{Saxe2018OnTI,pmlr-v97-goldfeld19a,Wu2019LearnabilityFT}. In this paper, we mainly focus on the optimization of the IB Lagrangian (Eq. \ref{eq:IB_lagrangian}) as well as the effect of the trade-off between information compression and prediction.
Previous studies have attempted to solve this problem by imposing some assumptions on the data distributions and introducing random noise into the hidden representations. \\

\noindent \textbf{VIB} \quad Alemi \cite{Alemi2017DeepVI} introduced VIB inspired by VAE \cite{Kingma2014AutoEncodingVB} to prediction tasks. By assuming $Z \sim \mathcal{N}(\mu|\sigma^2)$, they bound $\mathrm{MI}(X; Z)$ in the following equality,
\begin{equation*}
\mathrm{MI}(X; Z) \geq E_{X \sim Q(X)} \mathrm{KL}(P_{\theta}(Z|X)||P(Z))
\label{eq:vib}
\end{equation*}
They further replace the prediction term $\mathrm{MI}(Z; Y)$ by the cross-entropy loss. Subsequently, Matthew \emph{et al.} \cite{chalk2016relevant} improved VIB by introducing different surrogate marginal distributions (e.g., Student$^{'}$s t-distributions) on the intermediate representation layer. Achille and Soatto \cite{Achille2018InformationDL} approximated the IB Lagrangian by introducing a information dropout layer, which allows the network to selectively introduce multiplicative noise in the layer activation, and thus to control the flow of information.\\

\noindent \textbf{NIB} \quad Kolchinsky \cite{Kolchinsky2019NonlinearIB} introduced a novel non-parametric upper bound for mutual information by injecting noise to the representation layer and achieved the best trade-off between the information compression and prediction. Specifically, NIB introduced noise $\hat{Z} = Z + n(\sim \mathcal{N}(0|1))$ to approximate $\mathrm{MI}(X; Z)$, i.e.,
\begin{equation*}
\mathrm{MI}(X; Z) \approx \mathrm{MI}(X;\hat{Z}) \leq E_{i}\ln E_{j}e^{-\mathrm{KL}(p_{i}||p_{j})}
\label{eq:nib}
\end{equation*}
As a result, the learned intermediate representations form geometrically dense clusters. However, they failed to explore whether NIB can improve the performance on adversarial robustness.

The above IB methods introduce random noise to the intermediate representation to reduce its dependence on the input data, and mitigate adversarial effects \cite{Amjad2020LearningRF,Gu2015TowardsDN,Rakin2019ParametricNI}. However, these methods in fact failed to explain the role of the IB principle in their design.
In addition, selection of $\beta$ is critical to learn a meaningful representation since it controls the information compression and realize representation invariance. However, previous studies have set $\beta$ in an empirical range, which don't well capture the relationship between compression and prediction. 
Chechik \emph{et al.} \cite{Chechik2003InformationBF} noted the presence of trivial solution when $\beta \geq 1$, in which $\mathrm{P}(Z|X) = \mathrm{P}(Z)$ becomes the global minimum of the IB Lagrangian.
Wu \emph{et al.} \cite{Wu2019LearnabilityFT} introduced the concept of IB-Learnability, and showed that the IB Lagrangian will undergo a phase transition from the inability to learn to the ability to learn when varying $\beta$. Furthermore, they theoretically and experimentally provided the range of $\beta$ to avoid the failure of IB Lagrangian optimization. However, the effect of $\beta$ on learning invariant representation has still not been explored.

\subsection{Mutual Information Estimation}
Estimating the mutual information between the input data to a DNN and its hidden layers has long been a topic of research, with applications to representation learning \cite{Oord2018RepresentationLW} and deep learning \cite{ShwartzZiv2017OpeningTB,Saxe2018OnTI,pmlr-v97-goldfeld19a,Wu2019LearnabilityFT}. For example, Tishby \emph{et al.} \cite{ShwartzZiv2017OpeningTB} empirically suggested that trajectories in the information plane or learning processes of DNNs appeared to consist of two distinct phases: an initial ``fitting" phase where mutual information between the hidden layers and both the input and output increase, and a subsequent ``compression" phase where mutual information between the hidden layers and the input decreases. They claimed that this compression phase is responsible for the excellent generalization performance of DNNs. Saxe \emph{et al.} \cite{Saxe2018OnTI} argued that the ``compression" phase arises primarily due to the double-saturating tanh activation function utilized, which will disappear when replaced with the ReLU activation. In addition, Ziv \emph{et al.} \cite{pmlr-v97-goldfeld19a} observed that compression can occur in VIB and revealed that compression was driven by progressive geometric clustering of the representations of samples from the same class.

However, despite the strengths of mutual information and effectiveness of the above IB methods, mutual information is often impossible to compute analytically, and hard to estimate from high-dimensional samples. Most previous information-theoretic studies of deep learning \cite{ShwartzZiv2017OpeningTB, Saxe2018OnTI} approximate the mutual information by discretizing the outputs of neurons (i.e., the ``binning" operation). However, in practice DNN does not operate on the binned variables, but on the continuous ones. Moreover, there are many possible binning strategies, which yield different discrete random variables, and different mutual information with respect to the input. For example, Saxe \emph{et al.} \cite{Saxe2018OnTI} linked the compression phase occurred in \cite{ShwartzZiv2017OpeningTB} to saturation of neurons, which usually happens in layers with double-saturating activations (e.g., sigmoid) and could lead to the failure of binning strategy.

A variety of approaches for entropy and thereby mutual information estimation have been developed over the years, including $k$-nearest neighbors (kNN) techniques \cite{Lord2018GeometricKN} and kernel density estimation (KDE) techniques \cite{Kandasamy2015NonparametricVM, Han2017OptimalRO}. Wickstr{\o}m \emph{et al.} \cite{Wickstrm2019InformationPA} utilized a kernel tensor-based estimator of the Renyi's entropy, and provided the first comprehensive information plane analysis of contemporary large-scale DNNs and CNNs. Recently, trainable neural estimators \cite{Belghazi2018MutualIN} has also been utilized to estimate mutual information and the authors attempted to apply this technique to optimize IB principle. However, their focus is mainly on estimating mutual information and there is no implementation details and result analysis in their work. 
In this paper, we use the KDE technique \cite{Kandasamy2015NonparametricVM} to estimate the mutual information between input, labels and hidden layers, which is fast, effective and robust to the dimensionality.

\subsection{Adversarial Attacks}
Since the original work in \cite{Shamir2010LearningAG}, many different types of adversarial attacks have been proposed to fool a trained DNN through introducing barely visible perturbations upon input data. Several state-of-the-art white-box attacks will be investigated in this work, and are briefly introduced as follows.\\

\noindent \textbf{FGS Attack:}\quad  Fast Gradient Sign (FGS) method  \cite{Goodfellow2015ExplainingAH} hypothesizes that DNNs are vulnerable to adversarial perturbations because of their linear nature. It finds the adversarial examples by projecting the sample ($x$) along the direction of gradients with respect to the input. Specifically, $x$ is manipulated by adding or subtracting a small perturbation $\epsilon$ to each pixel. Whether adding or subtracting $\epsilon$ depends on whether the sign of the gradient for a pixel is positive or negative. Adding errors in the direction of the gradient means that the image is intentionally altered so that the model classification fails. The adversarial example can be formulated as,
\begin{equation*}
x_{\text{FGS}} = x + \epsilon \cdot{} \text{sign}(\bigtriangledown_x{}\mathcal{L}(\theta,x,y))
\end{equation*}
where $\bigtriangledown_x{}L$ is the gradient of the loss function with respect to the original input pixel vector $x$, $y$ is the true label vector for $x$ and  $\theta$ is the model hyperparameter vector.
If the perturbation is small, these adversarial examples are indistinguishable from normal examples to human, but the network performs significantly worse on them.\\

\noindent \textbf{TGS Attack:}\quad  Targeted Gradient Sign (TGS) attack was designed based on the FGS method. It attempts to encourage the model to misclassify samples in a specific way,
\begin{equation*}
x_{\text{TGS}} = x - \epsilon \cdot{} \text{sign}(\bigtriangledown_x{}\mathcal{L}(\theta,x,y_{\text{target}}))
\end{equation*}
where $y_{\text{target}}$ encodes an alternative set of labels we would like the model to predict instead. TGS can also be performed iteratively.
\\

\noindent \textbf{DeepFool Attack:} \quad DeepFool attack \cite{MoosaviDezfooli2016DeepFoolAS} is an iterative attack method which finds the minimal perturbation to cross the decision boundary based on the linearization of the classifier at each iteration. Given a sample $x_n$ belonging to the class $t$, the perturbation is formulated as,
\begin{equation*}
\begin{array}{ll}
\underset{\delta_n}{\operatorname{min}} & \left\|\delta_n\right\|_{p} \\
\text {s.t.} & \max _{j \neq t}\left\{g_{j}(x_n+\delta_n)\right\}-g_{t}(x_n+\delta_n) \geq 0
\end{array}
\end{equation*}
where $\delta_n$ is taken as the perturbation and $\left\| \cdot \right\|_{p}$ can be any norm specified by the user. The sum of all the perturbations from all the samples can be taken as the robustness performance of the given model. Note that $\delta_n$ can be taken as the margin, i.e., the distance from the sample $x_n$ to the classification boundary in a two-class linear classifier case. In non-linear and multi-label cases, the minimum attack distance will be iterated along the the vertical direction of the nearest classification boundary.

\section{Adversarial Information Bottleneck}\label{sec:AIB}
The difficulty of optimizing the IB Lagrangian lies in the compression term $\mathrm{MI}(X; Z)$, since the prediction term $\mathrm{MI}(Z; Y)$ follows \cite{Alemi2017DeepVI,Kolchinsky2019NonlinearIB},
\begin{equation}
\mathrm{MI}(Z; Y) \leq \mathrm{H}(Y) + E_{P(Z; Y)}\log P(Y|Z)
\end{equation}
where the second expectation term is equivalent to the usual cross-entropy loss in supervised learning, which can be optimized effectively in the settings of machine learning.

To optimize the compression term $\mathrm{MI}(X; Z)$, we recall that mutual information can be formulated as the Kullback-Leibler ($\mathrm{KL}$) divergence between the joint $P(X; Z)$ and the product of the marginals $P(X)P(Z)$. Recent studies utilized a dual formula to cast the estimation of $f$-divergences \cite{Nowozin2016fGANTG,Donsker1975AsymptoticEO,Belghazi2018MutualIN} (including the $\mathrm{KL}$-divergence) as part of an adversarial game between competing DNNs \cite{Nguyen2017DualDG}. Formally, the mutual information admits the following equality,
\begin{equation}
\mathrm{MI}(X; Z) = \max_{T:\Omega \rightarrow R}E_{P_{XZ}}T(x,z)-\log E_{P_{X}P_{Z}}e^{T(x,z)}
\label{eq:dual_kl}
\end{equation}
where the maximum is taken over all functions $T$ such that the two expectations are finite.\\

\noindent \textbf{Optimization}\quad In this paper, we use the neural network with an encoder-decoder architecture, where the encoder $G$ outputs the representation $Z$ by processing the input data $X$ and the decoder $D$ outputs the predicted labels based on $Z$. In addition, the encoder and decoder are parameterized by $\theta$ and $\phi$, respectively. Furthermore, we parameterize the function $T$ in Eq. \ref{eq:dual_kl} by a neural network parameterized by $\psi$ \cite{Belghazi2018MutualIN}. Therefore, the minimization of the compression term $\mathrm{MI}(X; Z)$ can be formulated as the following Min-Max problem,
\begin{equation*}
\min_{\theta}\max_{\psi}E_{p_{X,Z}}T_{\psi}(x,G_{\theta}(x))-\log E_{p_{X}p_{Z}}e^{T_{\psi}(x,G_{\theta}(x))}
\end{equation*}
where the inner maximization tends to acquire accurate estimates of mutual information by optimizing $T_{\psi}$, and the outer minimization on $G_{\theta}$ is optimized to compress the information flow of $G$.
\begin{figure}
  \centering
  \includegraphics[width=.47\textwidth]{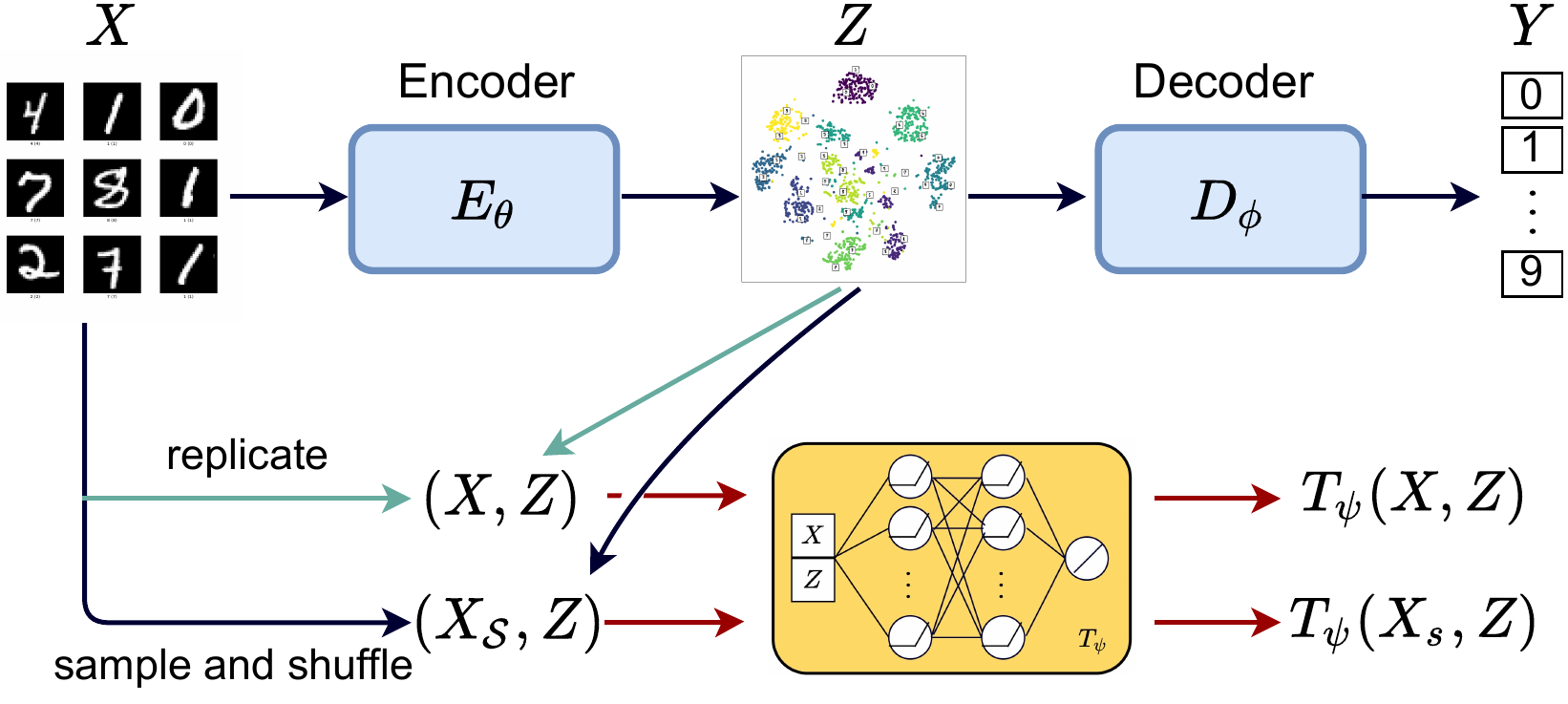}
  \caption{Illustration of AIB.}
  \label{img:structure}
\end{figure}
Optimizing Eq. (\ref{eq:dual_kl}) to converge after every gradient descent step guarantees us to stay on the desired manifold. This is an expensive procedure. Moreover, it may result in parameters that are far away from the ones obtained after the main gradient update.
We use two approximations to make the algorithm more efficient. First, we only do one step of descent on the function Eq. \ref{eq:dual_kl}. Secondly, instead of optimizing the mutual information between the whole data $X$ and representation $Z$, one can find a subset $\mathcal{S}$ of the most informative pixels and perform the update on them, denoted by $X_{\mathcal{S}}$. This procedure is meaningful since
\begin{equation}
\mathrm{MI}(X_{\mathcal{S}};Z) \leq \mathrm{MI}(X; Z)
\end{equation}
The two terms are equal if and only if $\mathcal{S}$ is the whole set of the pixels. Specifically, we determine $X_{\mathcal{S}}$ of pixels with large gradients, i.e., the gradients of $Z$ with respect to the inputs, since these selected pixels are the most informative ones about $Z$. The structure and algorithm of AIB are summarized in Fig. \ref{img:structure} and Algorithm \ref{alg:aib}, respectively.
\begin{algorithm}[htb]
\caption{Adversarial Information Bottleneck}    \label{alg:aib}
\begin{algorithmic}[1]
\Require  Datasets $\mathcal{D}=\{(x_i,y_i)\}, i=1,2,\cdots,N$; the number $k$ of steps to update $T_{\psi}$; the number $p$  of pixels sampled from the input data; the trade-off hyperparameter $\beta$.
\Ensure
\State \textbf{For} $k$ steps \textbf{do}
\State Sample two minibatch of $m$ samples $\{(x_i,y_i)\}_{i=1}^{m}$, $\{(\hat{x}_i,\hat{y}_i)\}_{i=1}^{m}$
\State Feed one minibatch samples into the decoder $G$ and obtain the representation $z_i = \{G_{\theta}(x_i)\}$
\State Select $p$ pixels with the most significant gradients w.r.t. $x_i$
\State Update $T_{\psi}$ by ascending its stochastic gradient:
\begin{equation*}
\nabla_{\psi} \frac{1}{m}\left[\sum_{i}^{m}T_{\psi}(x_i,z_i)-\log\frac{1}{m}\sum_{i}^{m}
e^{T_{\psi}(\hat{x}_{i},z_i)}\right]
\end{equation*}
\State \textbf{End for}
\State Update the encoder $G_{\theta}$ and decoder $D_{\phi}$ by descending  their respective stochastic gradient:
\begin{equation*}
\left\{
             \begin{array}{lr}
             \nabla_{\phi}\frac{1}{m}\sum_{i}^{m} y_{i} \log D_{\phi}(z_i), &  \\
             \nabla{\theta} \frac{1}{m}\sum_{i}^{m} y_{i} \log D_{\phi}(E_{\theta}(x_i)) + \beta[\frac{1}{m}\sum_{i}^{m}T_{\psi}(x_i,E_{\theta}(x_i))\\ \quad \quad \quad \quad \quad \quad \quad \quad \quad -\log \frac{1}{m}\sum_{i}^{m}e^{T_{\psi}(\hat{x}_{i},E_{\theta}(x_i))}], &
             \end{array}
\right.
\end{equation*}
\Output
\end{algorithmic}
\end{algorithm}

We name our proposed method as adversarial information bottleneck due to its ``adversarial" characteristic by solving a Min-Max optimization problem, which is very different from adversarial training by feeding the adversarial examples into the training process \cite{Goodfellow2015ExplainingAH}. In particular, the adversarial training required a computationally expensive inner loop in order to evaluate the adversarial perturbations, in which the cost becomes prohibitive with growing model complexity and input dimensionality.

\noindent \textbf{Hyperparameter Selection} \quad Another challenge in the optimization of the IB Lagrangian is that there is a lack of understanding about the relationship between $\beta$ and adversarial robustness. Specifically, the question under consideration is whether and how compression promotes better robustness by varying $\beta$? Furthermore, if so, how to select $\beta$ to achieve the best robustness?

Here, we provide a simple way to select $\beta$ based on an observation about the IB curve (Fig. 2). As $\beta$ decreases, the IB curve presents two distinct phases: (1) $\mathrm{MI}(X; Z)$ and $\mathrm{MI}(Z; Y)$ increase
simultaneously with the decrease of $\beta$; (2) once $\beta$ exceeds a critical value, $\mathrm{MI}(Z; Y)$ increases slowly or stay the same, while $\mathrm{MI}(X; Z)$ keeps increasing. Thus, the difference of the changing rate of the two values leads to a sharp knee and turning point in the IB curve.
We intuitively relate this knee point to the optimal hyperparameter which characterizes the best robustness. Numerical experiments clearly confirm this observation.
\begin{figure}
  \centering
  \includegraphics[width=.5\textwidth]{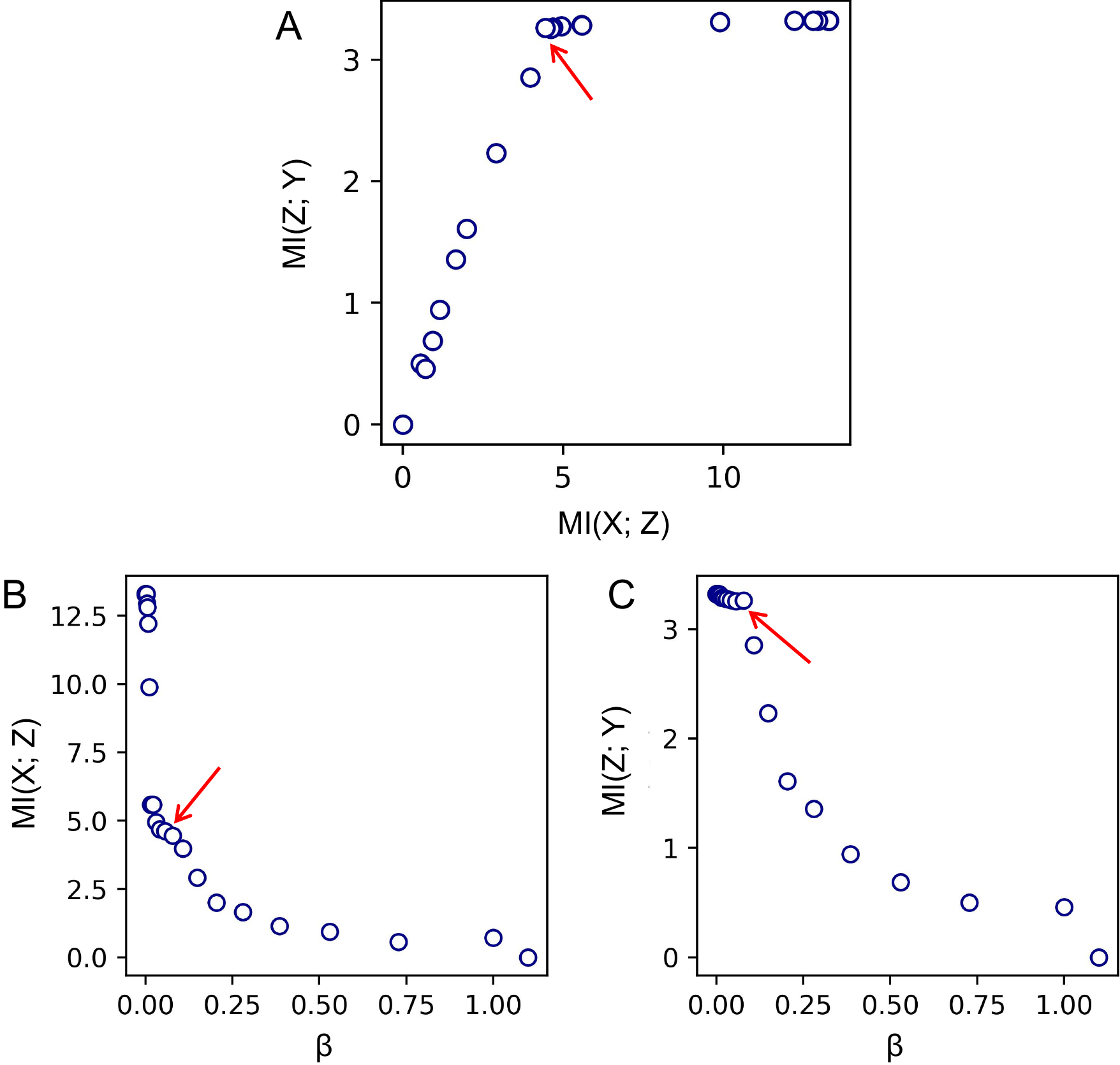}
  \caption{(A) Illustration of the IB curve of AIB on MNIST. (B) and (C) Plots of $\mathrm{MI}(X; Z)$ and $\mathrm{MI}(Z; Y)$ versus $\beta$, respectively. Each blue circle corresponds to a fully-converged model starting with independent initialization. The red arrows indicate the knee point determined in the IB curve in (A). }
  \label{img:fig2}
\end{figure}

\begin{figure*}
	\centering
	\includegraphics[width=.75\textwidth]{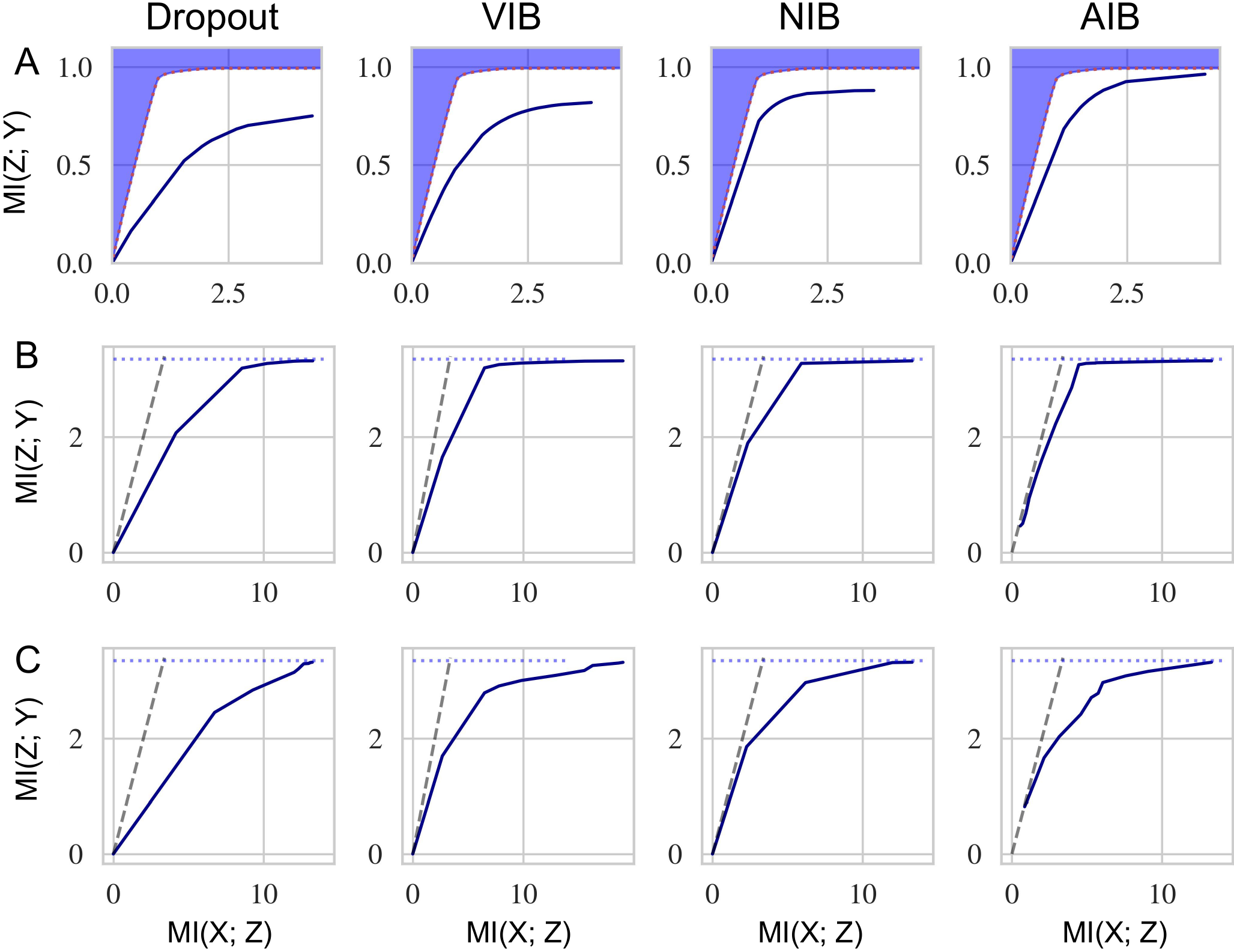}
	\caption{(A) The IB curves of all of the methods on the synthetic dataset. The red dashed line is the IB curve computed by the BA algorithm, the blue region indicates the non-achievable region in the information plane. (B) and (C) The IB curves on MNIST (B) and FashionMNIST (C), respectively.
		The black dashed line is the data-processing inequality bound $\mathrm{MI}(Z; Y) \leq \mathrm{MI}(X; Z)$, the blue dotted line indicates the value of $\mathrm{H}(Y)$ achieved by a baseline model trained only to optimize the cross-entropy loss. }
	\label{img:fig3}
\end{figure*}

\section{Experimental Results}\label{sec:experiments}
We first demonstrate the performance of AIB and the effectiveness of hyperparameter selection through exploring the IB curves. Next, we show AIB distinctly improves the adversarial robustness compared to other methods with three different adversarial attacks.

AIB is compared with VIB \cite{Alemi2017DeepVI} and NIB \cite{Kolchinsky2019NonlinearIB} on a synthetic dataset \cite{ShwartzZiv2017OpeningTB} and two common classification datasets MNIST \cite{LeCun1998GradientbasedLA} and FashionMNIST \cite{Xiao2017FashionMNISTAN} by varying hyperparameter $\beta$.  We adopt the typical DNNs trained with the cross-entropy loss to show its representation ability and robustness (denoted as ``Normal"). We also employ the popular ``dropout" method (denoted as Dropout) with dropout rate as its regularization hyperparameter for evaluation. We still call the trade-off curve relating to ``dropout" as the IB curve for convenience.

\subsection{Information Compression and Extraction}
\noindent \textbf{Synthetic Experiment.}\quad First, we utilize a synthetic dataset as suggested in \cite{ShwartzZiv2017OpeningTB} and find the bottleneck variable through the Blahut-Arimoto (BA) algorithm \cite{Tishby2000TheIB}, which can be taken as the golden standard solution. The synthetic dataset is constructed as binary decision rules which are invariant under $O(3)$ rotations of the sphere, with 12 binary inputs that represent 12 uniformly distributed points on a two-dimensional sphere. Therefore, there are 4096 different patterns in total and the classification rule is designed such that $p(y=0)=p(y=1)\simeq 0.5$ and the mutual information $\mathrm{MI}(X; Y) \simeq 0.99$ bits. With the BA algorithm, minimization of Eq. \ref{eq:IB_lagrangian} is performed by the converging alternating iterations as follows, \\
\begin{equation}
\left\{
\begin{array}{lr}
p_{k}(z|x) = \frac{p_{k}(z)}{S_{k}(x,\beta)}\exp(-\beta d(x,z))  &\\
p_{k+1}(z) = \Sigma_{x}p(x)p_{k}(z|x)  &\\
p_{k+1}(y|z) = \Sigma_{y}p(y|x)p_{k}(x|z) &\\
\end{array}
\right.
\end{equation}
where $S_{k}(x,\beta)$ is the normalization function, $z$ is one of the ten discrete states of $Z$, and $k$ denotes the iteration step. Following the above formula, the optimal IB bound can be found by varying the trade-off hyperparameter $\beta$. The region above the IB bound is non-achievable and the approximation closeness to the theoretical bound demonstrates the effectiveness of information extraction (Fig. \ref{img:fig3}A).

A fully connected feed-forward neural network (12-10-10-2-10-2) with no other architecture constraints is utilized to evaluate all the methods. The ReLU function is adopted for activation, and the Softmax function is employed in the final layer. The networks were trained using the cross-entropy loss function and the optimizer Adam with learning rate 0.0002. For easier estimation of mutual information, we set the outputs of the two-node hidden layer as the bottleneck variable $Z$. The IB curves are plotted by calculating the mutual information between the input data, labels and the bottleneck layer $Z$, respectively. We note that the network capacity is relatively small, so the IB curves explored by these methods are not very close to the optimal solution.

\noindent \textbf{Real-world Experiment.}\quad Two common real-world classification datasets (MNIST and FashionMNIST) are utilized to evaluate the performance of AIB. Each image of MNIST and FashionMNIST consists of 28-by-28 pixels (784 total pixels), i.e., $X \in R^{784}$, and is classified into one of ten classes corresponding to the digit or clothing identity (Y $\in$ \{0, $\cdots$, 9\}). Both the two datasets contain 60,000 training images and 10,000 testing images. 
MNIST is a relative simple dataset for modern machine learning and FashionMNIST is a more complex dataset.
Note that the BA algorithm is not applicable here since the dimension of these two datasets is pretty high and the conditional probability is incalculable. But the IB curves are still bounded according to the DPI (Eq. \ref{eq:DPI}) and definition of mutual information,
\begin{equation*}
\mathrm{H}(Y) \geq \mathrm{MI}(Z; Y) \leq \mathrm{MI}(X; Z)
\label{eq:ib_bound}
\end{equation*}

Here we use the similar setting with the synthetic experiment except the network architecture. We apply a simple fully connected network and a convolutional neural network onto MNIST and FashionMNIST, respectively, since the latter is a more difficult task. The fully connected layers have units 784-128-128-10-128-10 and the 10-unit layer is set as the bottleneck layer. The simple convolutional neural network consists of 3 $\times$ 3 $\times$ 32 and 3$\times$ 3 $\times$ 64 convolutional layers followed by a 2 $\times$ 2 max pooling and 2 fully connected layers (64-128). The 64-unit layer is set as the bottleneck layer. In both the experiments, the training images are divided into training and validation set at a ratio of 4:1. We pick the model that performs the best on clean accuracy of the validation set. 

We present the IB curves of all the methods on the synthetic and real-world datasets respectively (Fig. \ref{img:fig3}). Note that all the IB curves appear smooth and convex, which start from the point (0,0) corresponding to a large $\beta$ and stop at a point corresponding to $\beta = 0$, respectively. We come to the following conclusions,
\begin{itemize}
\item On the synthetic dataset, AIB performs the best compared to Dropout, VIB and NIB, and its IB curve is mostly close to the bound determined by the BA algorithm.
\item On all the datasets, the IB methods (AIB, NIB and VIB) perform better than Dropout in terms of the nuisance information compression. They achieve better prediction values at the same level of compression. Particularly, the performance of AIB is better (Synthetic and MNIST) or comparable (FashionMNIST) compared with that of NIB and VIB.
\item The curves of the IB methods present a phase transition phenomenon to some extent: (1) the mutual information between the representation and the input data $\mathrm{MI}(X; Z)$, labels $\mathrm{MI}(Z; Y)$ increases simultaneously; (2) once $\beta$ is less than a critical value, $\mathrm{MI}(Z; Y)$ increases slowly or stay the same level, while $\mathrm{MI}(X; Z)$ keeps increasing. 
\end{itemize}

\begin{figure}
	\centering
	\includegraphics[width=.48\textwidth]{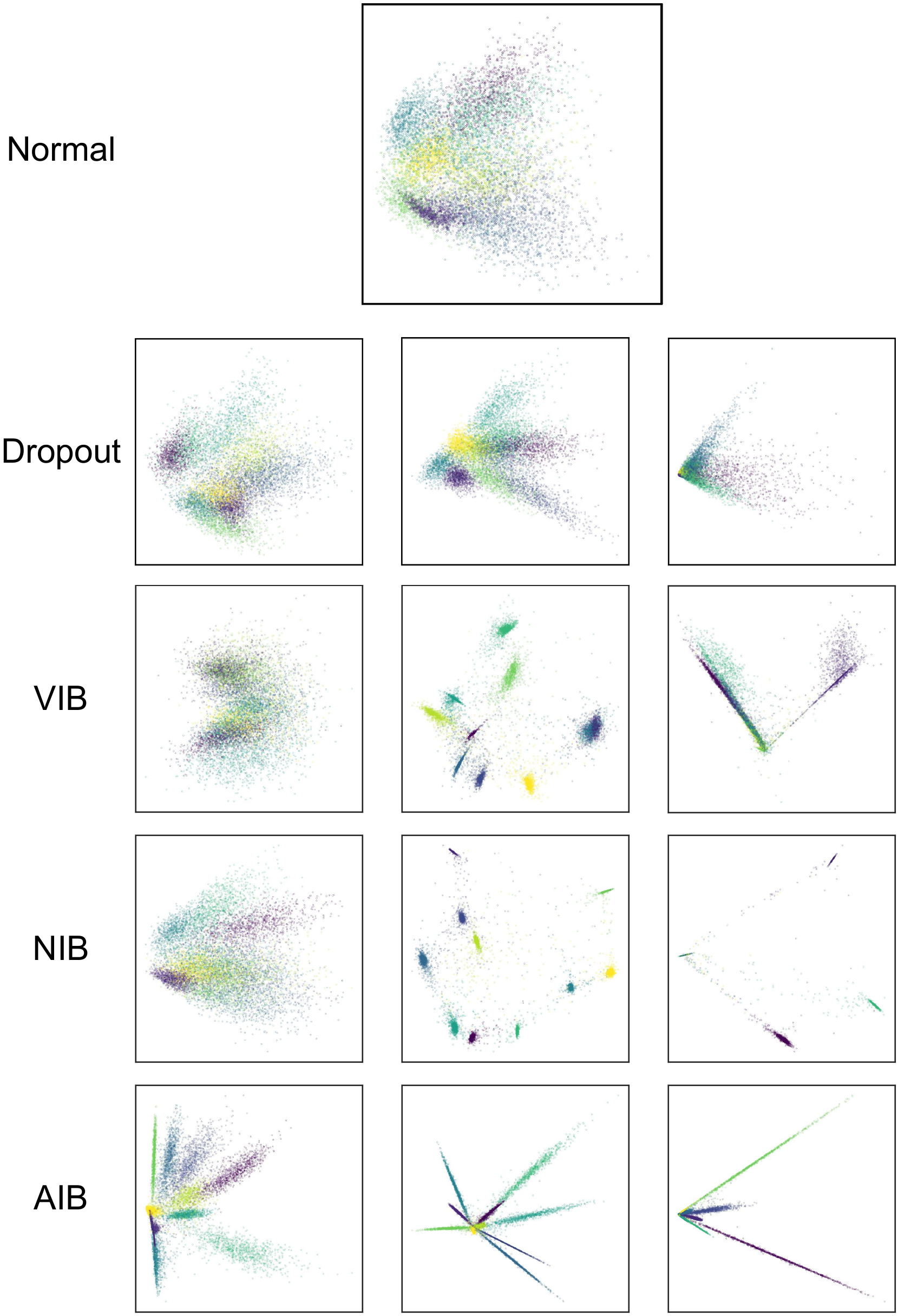}
	\caption{Projection of the bottleneck layer variable for the methods on MNIST with the increase of $\beta$ (from left to right) using principal component analysis. The middle corresponds to the hyperparameter of the knee point of the IB curve. Note the Normal method has no regularization hyperparameter.}
	\label{img:fig4}
\end{figure}

\noindent \textbf{Compression Promotes Geometric Representation}\quad First, we consider the clustering performance of all the methods on the bottleneck layer. We plot the visualization of the two-dimensional projections for them on MNIST (Fig. \ref{img:fig4}). 
Training with the IB Lagrangian or cross-entry enables different digits to fall into well-separated clusters, in which different architectures lead to clusters with diverse shapes. Specifically, the clusters of AIB tend to be line-shaped, indicating it has the largest decision margin. In addition, the geometric representation of AIB is significantly better than the regular Dropout method. This is consistent with the information compression results (Fig. \ref{img:fig3}) since the bits of compact clusters is smaller.

Note that Fig. \ref{img:fig4} (\textbf{middle}) corresponds to $\beta$ on the knee point of the IB curves (see next Section 4.2). Then their respective geometric representations get more compact (\textbf{middle}) compared to others (\textbf{left} and \textbf{right}). This observation gives us an promising way for selecting hyperparameter, which provide the best invariance and robustness against adversarial perturbations.\\


\begin{figure}
  \centering
  \includegraphics[width=.49\textwidth]{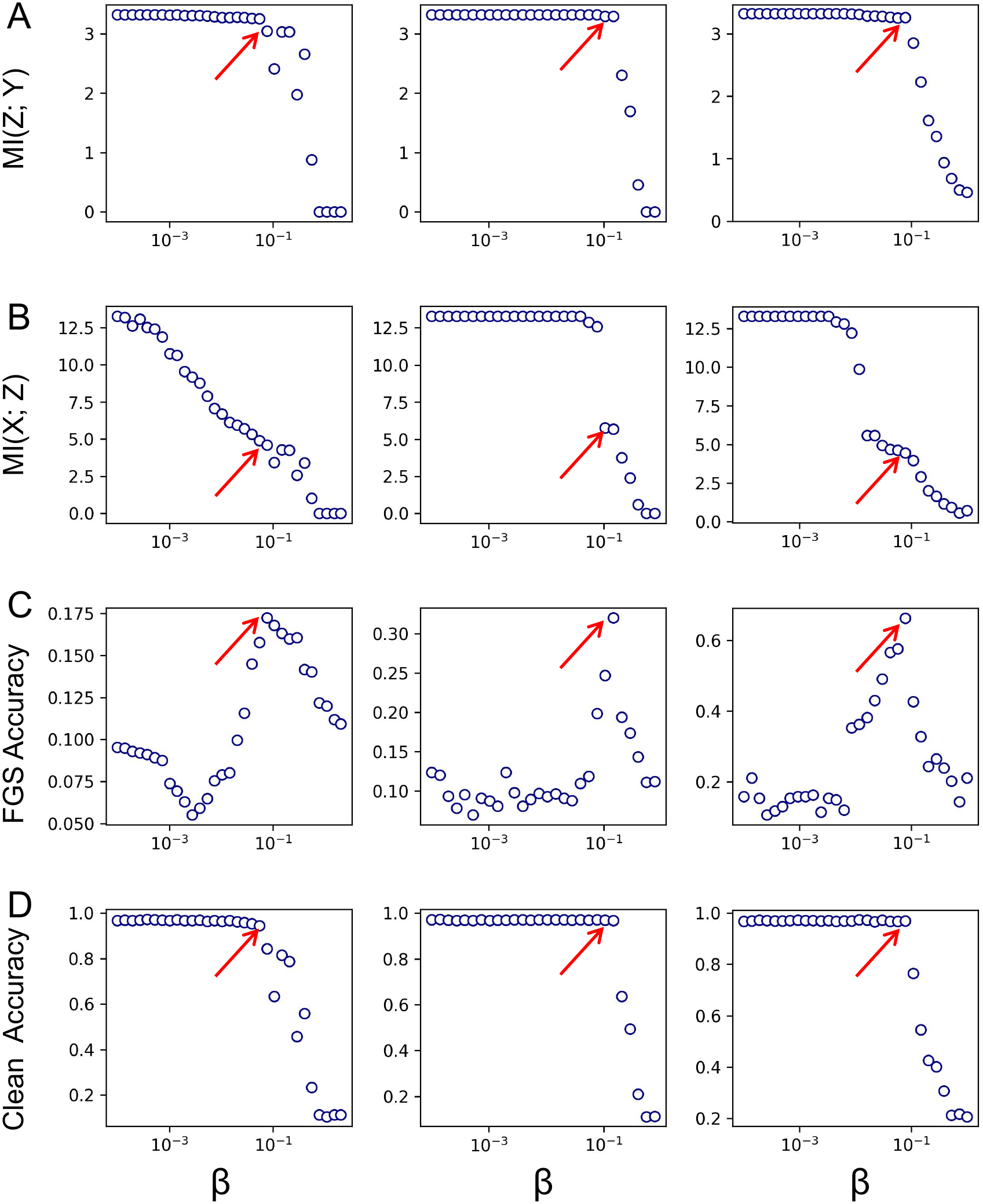}
  \caption{(A) and (B) $\mathrm{MI}(Z; Y)$ and $\mathrm{MI}(X; Z)$ evaluated on MNIST of the IB methods trained with their corresponding trade-off hyperparameter $\beta$. 
  (C) Accuracy on MNIST generated by FGS attack of the IB methods in terms of the trade-off hyperparameter $\beta$ (perturbation $\epsilon = 0.1$). (D) Clean accuracy on MNIST. Each blue circle corresponds to a fully-converged model starting with independent initialization. The $x$-axis is scaled differently for better visualization. The red arrows indicate the $y$-axis value trained with the hyperparameter corresponding to the knee point of the IB curve.}
  \label{img:fig5}
\end{figure}

\subsection{Adversarial Robustness and Defense}

In this section, we evaluate the robustness of all the models against the FGS, TGS and Deepfool attacks on MNIST, FashionMNIST as well as CIFAR-10 \cite{Krizhevsky2009LearningML}. The CIFAR-10 dataset is an established computer-vision dataset used for object recognition, and consists of 60,000 32x32 color images belonging to 10 object classes. All the images are divided into training, validation and test sets at a ratio of 4:1:1. We use a complex convolutional neural network on CIFAR-10. All the methods are trained using the optimizer Adam with learning rate 0.0002 and batch size 256. The methods are stopped if they don't improve the accuracy on validation set within 20 epochs. 

For the FGS and TGS attacks, we test all the methods against the adversarial examples generated for each strategy with different perturbation ranging from 0.05 to 0.3. Particularly, for the TGS attack, we pick the targets via increasing the original labels $y$ by 1 (modulo 10).
For the Deepfool attack, we compute the minimum perturbation upon all of the test samples and take $||\boldsymbol{\delta}||_2 = \sqrt{\Sigma \delta_n^2}$ as the evaluation metric to measure the robustness of DNN, where a greater value of $||\delta||_{2}$ normally indicates a DNN possesses higher robustness against potential adversarial attacks.
We use cleverhans library \cite{papernot2018cleverhans} to generate all the adversarial examples.

We first select the hyperparameter in a reasonable range in which the method has a pretty high clean accuracy. This is critical since the model with bad clean accuracy may present better robustness (Fig. \ref{img:fig5}C and D). 
We show the changing trend of $\mathrm{MI}(Z; Y)$,  $\mathrm{MI}(X; Z)$, accuracy on the adversarial examples generated by FGS attack (perturbation $\epsilon = 0.1$) and clean accuracy in terms of $\beta$ (Fig. \ref{img:fig5}). First, as the trade-off $\beta$ increases, the clean accuracy stays almost the same and adversarial accuracy increases distinctly until it reaches a critical point, which precisely corresponds to $\beta$ of the knee point in the IB curves. 
Secondly, once the trade-off $\beta$ exceeds the knee point, the clean accuracy and adversarial accuracy decease simultaneously. As a result, we select the $\beta$ ranging from 0 to that corresponding to the knee point in the IB curve.

Table \ref{table:fgs} reports the results of all the adversarial robustness to the FGS and TGS attacks on MNIST, FashionMNIST and CIFAR-10 at a variety of perturbation strengths $\epsilon$. 
The performance of VIB is comparable to Dropout method in a range of small perturbations. Although the performance of NIB on MNIST is significantly better than VIB and Dropout, it is far less impressive on more complex datasets (FashionMNIST and CIFAR-10). However, the performance of AIB is the best across all the datasets and attacks with different perturbations.

\begin{table*}[]
\centering
\caption{Performance comparison of clean- and perturbed-data (under FGS and TGS attacks) on MNIST in terms of accuracy (mean$\pm$std), FashionMNIST and CIFAR-10.}
\setlength{\tabcolsep}{3.0mm}{
\begin{tabular}{clllllll}
\toprule[1pt]
\multicolumn{1}{l}{}                          &                      &              & \multicolumn{5}{c}{Methods}                         \\ \cmidrule[0.5pt]{4-8}
\multicolumn{1}{l}{Datasets}                  & Attacks              & $\epsilon$ & \multicolumn{1}{c}{Normal} & \multicolumn{1}{c}{Dropout} & \multicolumn{1}{c}{VIB}   & \multicolumn{1}{c}{NIB}   & \multicolumn{1}{c}{AIB}            \\ \midrule[0.5pt]
\multirow{7}{*}{MNIST}                        & clean                & 0            & 0.970$\pm$0.005  & 0.974$\pm$0.004  & 0.971$\pm$0.004 & 0.973$\pm$0.002& \textbf{0.975}$\pm$\textbf{0.005} \\
                                              & \multirow{3}{*}{FGS} & 0.05         & 0.405$\pm$0.086  & 0.638$\pm$0.102   & 0.454$\pm$0.080 & 0.721$\pm$0.055 & \textbf{0.895}$\pm$\textbf{0.110} \\
                                              &                      & 0.10         & 0.079$\pm$0.010  & 0.361$\pm$0.066   & 0.164$\pm$0.025 & 0.451$\pm$0.016 & \textbf{0.721}$\pm$\textbf{0.046} \\
                                              &                      & 0.15         & 0.036$\pm$0.006  & 0.310$\pm$0.073   & 0.120$\pm$0.011 & 0.293$\pm$0.043 & \textbf{0.464}$\pm$\textbf{0.074} \\
                                              & \multirow{3}{*}{TGS} & 0.1          & 0.094$\pm$0.001 & 0.275$\pm$0.028   & 0.191$\pm$0.015 & 0.481$\pm$0.082 & \textbf{0.825}$\pm$\textbf{0.120} \\
                                              &                      & 0.2          & 0.002$\pm$0.000 & 0.268$\pm$0.010   & 0.165$\pm$0.008 & 0.215$\pm$0.014 & \textbf{0.468}$\pm$\textbf{0.064} \\
                                              &                      & 0.3          & 0.001$\pm$0.000 & 0.263$\pm$0.012   & 0.151$\pm$0.010 & 0.135$\pm$0.009 & \textbf{0.286}$\pm$\textbf{0.023} \\ \midrule[0.5pt]
\multirow{7}{*}{Fashion}                      & clean                & 0            & 0.902$\pm$0.005  & \textbf{0.909}$\pm$\textbf{0.006}   & 0.903$\pm$0.010 & 0.902$\pm$0.007 & 0.902$\pm$0.006 \\
                                              & \multirow{3}{*}{FGS} & 0.05         & 0.393$\pm$0.045  & 0.623$\pm$0.132   & 0.452$\pm$0.068 & 0.475$\pm$0.076 & \textbf{0.635}$\pm$\textbf{0.044} \\
                                              &                      & 0.10         & 0.177$\pm$0.012  & 0.430$\pm$0.054   & 0.171$\pm$0.032 & 0.218$\pm$0.019 & \textbf{0.466}$\pm$\textbf{0.057} \\
                                              &                      & 0.15         & 0.084$\pm$0.006  & \textbf{0.397}$\pm$\textbf{0.025}   & 0.101$\pm$0.011 & 0.163$\pm$0.016 & 0.320$\pm$0.015 \\
                                              & \multirow{3}{*}{TGS} & 0.1          & 0.341$\pm$0.038  & 0.563$\pm$0.050   & 0.284$\pm$0.024 & 0.424$\pm$0.056 & \textbf{0.576}$\pm$\textbf{0.056} \\
                                              &                      & 0.2          & 0.064$\pm$0.010  & 0.368$\pm$0.036   & 0.114$\pm$0.010 & 0.155$\pm$0.016 & \textbf{0.461}$\pm$\textbf{0.023} \\
                                              &                      & 0.3          & 0.018$\pm$0.002  & 0.333$\pm$0.018   & 0.106$\pm$0.008 & 0.146$\pm$0.010 & \textbf{0.387}$\pm$\textbf{0.012} \\ \midrule[0.5pt]

\multicolumn{1}{l}{\multirow{7}{*}{CIFAR-10}} & clean                & 0            & 0.851$\pm$0.028  & \textbf{0.860}$\pm$\textbf{0.022}   & 0.853$\pm$0.025 & 0.854$\pm$0.032 & 0.850$\pm$0.026 \\
                                              & \multirow{3}{*}{FGS} & 0.05         & 0.251$\pm$0.031  & \textbf{0.465}$\pm$\textbf{0.120}   & 0.387$\pm$0.047 & 0.366$\pm$0.068 & 0.418$\pm$0.076 \\
                                              &                      & 0.10         & 0.060$\pm$0.001  & 0.145$\pm$0.016   & 0.160$\pm$0.005 & 0.225$\pm$0.021 & \textbf{0.230}$\pm$\textbf{0.031} \\
                                              &                      & 0.15         & 0.020$\pm$0.000  & 0.104$\pm$0.002   & 0.074$\pm$0.001 & 0.113$\pm$0.005 & \textbf{0.120}$\pm$\textbf{0.005} \\
                                              & \multirow{3}{*}{TGS} & 0.1          & 0.220$\pm$0.026  & 0.246$\pm$0.045   & 0.270$\pm$0.016 & 0.319$\pm$0.043 & \textbf{0.350}$\pm$\textbf{0.032} \\
                                              &                      & 0.2          & 0.086$\pm$0.002  & 0.101$\pm$0.000   & 0.110$\pm$0.001 & 0.124$\pm$0.006 & \textbf{0.125}$\pm$\textbf{0.008} \\
                                              &                      & 0.3          & 0.032$\pm$0.000  & 0.010$\pm$0.000   & 0.086$\pm$0.002 & 0.105$\pm$0.008 & \textbf{0.109}$\pm$\textbf{0.006}               \\
\bottomrule[1pt]
\end{tabular}}\label{table:fgs}
\end{table*}

\begin{figure}
  \centering
  \includegraphics[width=.49\textwidth]{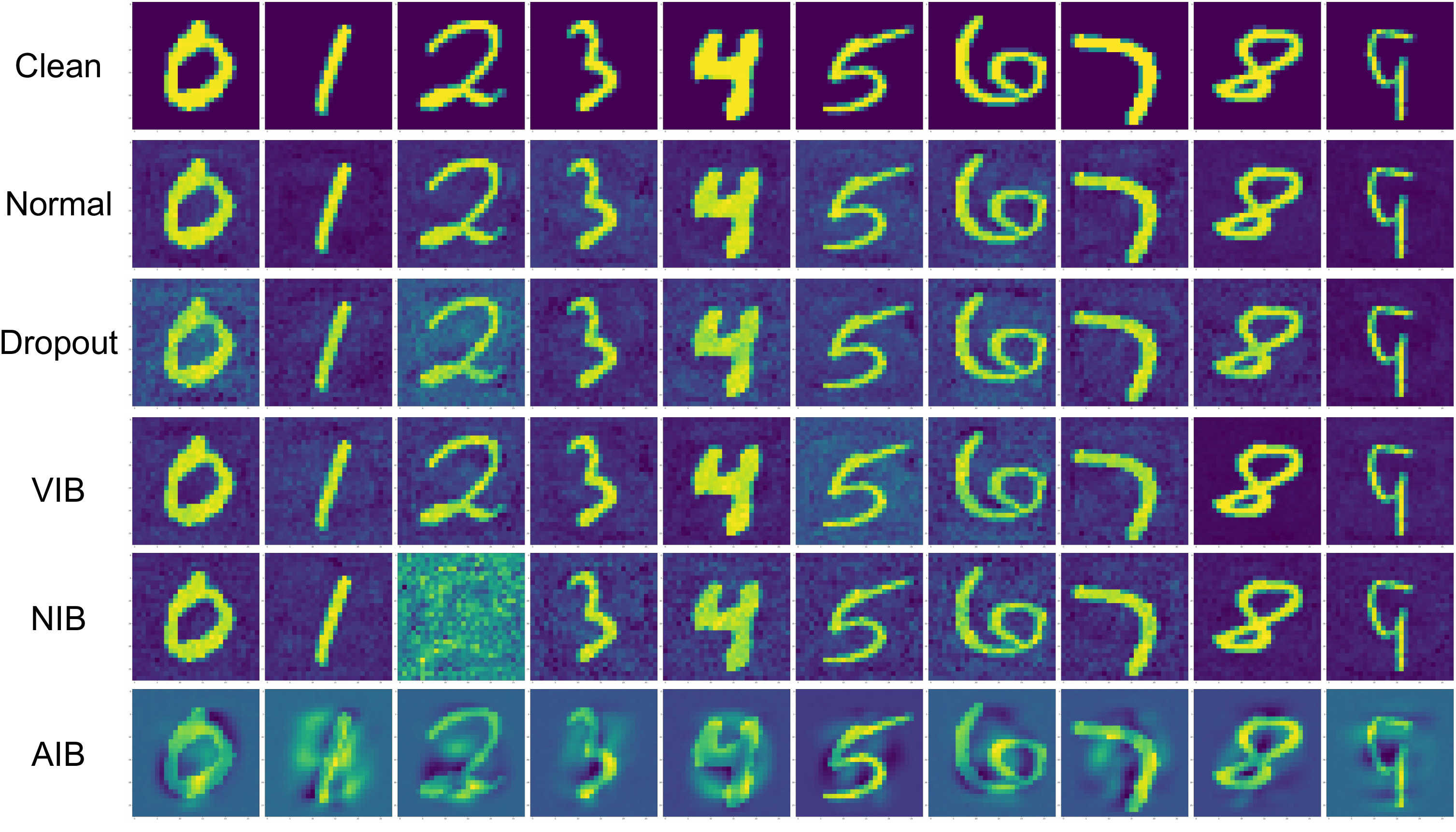}
  \caption{Adversarial examples of the Deepfool attack to all of the methods on different digits of MNIST. The blurriness of the picture indicates the robustness of the model to adversarial perturbations.}
  \label{img:fig6}
\end{figure}

It should be noted that only improving robustness against FGS and TGS attacks does not necessarily mean better accuracy against other attacks. Typically, Deepfool attack with $L_2$ norm could reach 100$\%$ success rate against any defense methods. Thus, the total perturbations required to fool a network gives more message about its robustness in general, and the large perturbations mean better robustness.
Table \ref{table:deepfool} presents the performance of all the methods against the DeepFool attack on MNIST, FashionMNIST and CIFAR-10. AIB consistently performs the best in all the datasets, while VIB and NIB only achieve similar performance with the Dropout method. In addition, Fig. \ref{img:fig6} demonstrates the adversarial examples generated by Deepfool attack applying to all the models on different digits of MNIST. As a result, the adversarial examples generated on AIB are the most unrecognizable ones, indicating that the examples need to move farther to fool it.

\begin{table}[]
\centering
\caption{Performance comparison of Deepfool attack with $L_2$ norm (mean$\pm$std) on MNIST, FashionMNIST and CIFAR-10.}
\setlength{\tabcolsep}{3mm}{
\begin{tabular}{llll}
\toprule[1pt]
                                                    & \multicolumn{3}{c}{Datasets}                                                                 \\
\cmidrule[0.5pt]{2-4}
Methods                                               & \multicolumn{1}{c}{MNIST}                         & \multicolumn{1}{c}{Fashion}                       & \multicolumn{1}{c}{CIFAR-10}                       \\ \cmidrule[0.5pt]{1-4}
Normal                                              & 1.130$\pm$0.011 & 0.800$\pm$0.009 &  0.502$\pm$0.015\\
Dropout                                             & 1.862$\pm$0.015 & 1.271$\pm$0.012 & 1.022$\pm$0.040 \\
VIB                                      & 1.360$\pm$0.015 & 1.125$\pm$0.022 & 1.114$\pm$0.035 \\
NIB                                        & 1.264$\pm$0.031 & 1.141$\pm$0.025 &  1.125$\pm$0.031\\
AIB                                                 &                              \textbf{2.204}$\pm$\textbf{0.042}  &   \textbf{1.651}$\pm$\textbf{0.030}                       &  \textbf{1.401}$\pm$\textbf{0.043}                                                   \\ \bottomrule[1pt]
\end{tabular}}\label{table:deepfool}
\end{table}


\section{Conclusion}
In this paper, we propose adversarial information bottleneck (AIB) to optimize the IB Lagrangian by introducing an adversarial regularization term to approximate the information compression term $\mathrm{MI}(X;Z)$.
AIB learns compact or even line-shaped representation and compress more nuisance information given the same prediction accuracy. Experimental results demonstrate that AIB performs the best in resisting adversarial perturbations compared to VIB, NIB and even others.


We further conduct a detailed and in-depth exploration about the effects of $\beta$ on learning invariant representations and mitigating adversarial perturbations. As a result, we empirically demonstrate that the hyperparameter corresponding to the knee point of the IB curve performs the best in geometric representation and adversarial robustness. We emphasize that this can help to understand the effect of compression on model robustness and hyperparameter selection.

In the future, it will be an interesting topic to learn invariant network weights and gradients with the proposed adversarial compression term which does not require an analytical form about their distributions. Furthermore, we claim that the information-theoretical perspective on the selection of trade-off $\beta$ is not only applicable to the IB models, but other defense models with a hyperparameter specifying the penalty strength. 

\ifCLASSOPTIONcompsoc
  \section*{Acknowledgments}
\else
  \section*{Acknowledgment}
\fi
The authors would like to thank Chihao Zhang, Kuo Gai and Rui Zhang for their helpful discussion. This work has been partially supported by the National Key R\&{D} Program of China [2019YFA0709501]; the National Natural Science Foundation of China [61621003]; National Ten Thousand Talent Program for Young Top-notch Talents; CAS Frontier Science Research Key Project for Top Young Scientist [QYZDB-SSW-SYS008].

\ifCLASSOPTIONcaptionsoff
  \newpage
\fi


\bibliographystyle{plain}
\bibliography{reference.bib}

\begin{thebibliography}{10}
\providecommand{\url}[1]{#1}
\csname url@samestyle\endcsname
\providecommand{\newblock}{\relax}
\providecommand{\bibinfo}[2]{#2}
\providecommand{\BIBentrySTDinterwordspacing}{\spaceskip=0pt\relax}
\providecommand{\BIBentryALTinterwordstretchfactor}{4}
\providecommand{\BIBentryALTinterwordspacing}{\spaceskip=\fontdimen2\font plus
\BIBentryALTinterwordstretchfactor\fontdimen3\font minus
  \fontdimen4\font\relax}
\providecommand{\BIBforeignlanguage}[2]{{%
\expandafter\ifx\csname l@#1\endcsname\relax
\typeout{** WARNING: IEEEtran.bst: No hyphenation pattern has been}%
\typeout{** loaded for the language `#1'. Using the pattern for}%
\typeout{** the default language instead.}%
\else
\language=\csname l@#1\endcsname
\fi
#2}}
\providecommand{\BIBdecl}{\relax}
\BIBdecl

\bibitem{He2016DeepRL}
K.~He, X.~Zhang, S.~Ren, and J.~Sun, ``Deep residual learning for image
  recognition,'' \emph{2016 IEEE Conference on Computer Vision and Pattern
  Recognition (CVPR)}, pp. 770--778, 2016.

\bibitem{Huang2017DenselyCC}
G.~Huang, Z.~Liu, and K.~Q. Weinberger, ``Densely connected convolutional
  networks,'' \emph{2017 IEEE Conference on Computer Vision and Pattern
  Recognition (CVPR)}, pp. 2261--2269, 2017.

\bibitem{AbdelHamid2014ConvolutionalNN}
O.~Abdel-Hamid, A.~rahman Mohamed, H.~Jiang, L.~Deng, G.~Penn, and D.~Yu,
  ``Convolutional neural networks for speech recognition,'' \emph{IEEE/ACM
  Transactions on Audio, Speech, and Language Processing}, vol.~22, pp.
  1533--1545, 2014.

\bibitem{Silver2017MasteringCA}
D.~Silver, T.~Hubert, J.~Schrittwieser, I.~Antonoglou, M.~Lai, A.~Guez,
  M.~Lanctot, L.~Sifre, D.~Kumaran, and e.~Graepel, Thore, ``Mastering chess
  and shogi by self-play with a general reinforcement learning algorithm,''
  \emph{ArXiv}, vol. abs/1712.01815, 2017.

\bibitem{Cao2019SimpleTO}
Z.~Cao and S.~Zhang, ``Simple tricks of convolutional neural network
  architectures improve dna-protein binding prediction,''
  \emph{Bioinformatics}, vol. 35 11, pp. 1837--1843, 2019.

\bibitem{Krueger2017DeepND}
D.~Krueger, N.~Ballas, S.~Jastrzebski, D.~Arpit, S.~Kanwal, T.~Maharaj,
  E.~Bengio, A.~Fischer, and A.~Courville, ``Deep nets don't learn via
  memorization,'' in \emph{ICLR}, 2017.

\bibitem{Bashir2020AnIP}
D.~Bashir, G.~D. Montanez, S.~Sehra, P.~S. Segura, and J.~Lauw, ``An
  information-theoretic perspective on overfitting and underfitting,''
  \emph{ArXiv}, vol. abs/2010.06076, 2020.

\bibitem{10.1093/nsr/nwaa159}
Z.~Zhang and S.~Zhang, ``{Towards understanding residual and dilated dense
  neural networks via convolutional sparse coding},'' \emph{National Science
  Review}, 2020, nwaa159.

\bibitem{Ribeiro2016WhySI}
M.~T. Ribeiro, S.~Singh, and C.~Guestrin, ``"why should i trust you?":
  Explaining the predictions of any classifier,'' \emph{Proceedings of the 22nd
  ACM SIGKDD International Conference on Knowledge Discovery and Data Mining},
  2016.

\bibitem{ShwartzZiv2017OpeningTB}
R.~Shwartz-Ziv and N.~Tishby, ``Opening the black box of deep neural networks
  via information,'' \emph{ArXiv}, vol. abs/1703.00810, 2017.

\bibitem{Tishby2015DeepLA}
N.~Tishby and N.~Zaslavsky, ``Deep learning and the information bottleneck
  principle,'' \emph{2015 IEEE Information Theory Workshop (ITW)}, pp. 1--5,
  2015.

\bibitem{GiladBachrach2003AnIT}
R.~Gilad-Bachrach, A.~Navot, and N.~Tishby, ``An information theoretic tradeoff
  between complexity and accuracy,'' \emph{Lecture Notes in Computer Science},
  pp. 595--609, 07 2003.

\bibitem{Shamir2010LearningAG}
O.~Shamir, S.~Sabato, and N.~Tishby, ``Learning and generalization with the
  information bottleneck,'' \emph{Theoretical Computer Science}, vol. 411,
  no.~29, pp. 2696 -- 2711, 2010, algorithmic Learning Theory (ALT 2008).

\bibitem{Bassily2018LearnersTU}
R.~Bassily, S.~Moran, I.~Nachum, J.~Shafer, and A.~Yehudayoff, ``Learners that
  use little information,'' in \emph{Proceedings of Algorithmic Learning
  Theory}, ser. Proceedings of Machine Learning Research, F.~Janoos, M.~Mohri,
  and K.~Sridharan, Eds., vol.~83.\hskip 1em plus 0.5em minus 0.4em\relax PMLR,
  07--09 Apr 2018, pp. 25--55.

\bibitem{Tishby2000TheIB}
N.~Tishby, F.~C. Pereira, and W.~Bialek, ``The information bottleneck method,''
  \emph{ArXiv}, vol. physics/0004057, 2000.

\bibitem{phdthesis}
B.~Rodriguez~Gálvez, ``The information bottleneck: Connections to other
  problems, learning and exploration of the ib curve,'' Ph.D. dissertation, 06
  2019.

\bibitem{Slonim2000DocumentCU}
N.~Slonim and N.~Tishby, ``Document clustering using word clusters via the
  information bottleneck method,'' in \emph{SIGIR '00}, 2000.

\bibitem{Goldberger2002UnsupervisedIC}
J.~Goldberger, H.~Greenspan, and S.~Gordon, ``Unsupervised image clustering
  using the information bottleneck method,'' in \emph{DAGM-Symposium}, 2002.

\bibitem{pmlr-v97-goldfeld19a}
Z.~Goldfeld, E.~Van Den~Berg, K.~Greenewald, I.~Melnyk, N.~Nguyen,
  B.~Kingsbury, and Y.~Polyanskiy, ``Estimating information flow in deep neural
  networks,'' ser. Proceedings of Machine Learning Research, K.~Chaudhuri and
  R.~Salakhutdinov, Eds., vol.~97.\hskip 1em plus 0.5em minus 0.4em\relax Long
  Beach, California, USA: PMLR, 09--15 Jun 2019, pp. 2299--2308.

\bibitem{Saxe2018OnTI}
A.~M. Saxe, Y.~Bansal1, J.~Dapello, M.~Advani, A.~Kolchinsky, B.~D. Tracey, and
  D.~D. Cox, ``On the information bottleneck theory of deep learning,'' in
  \emph{ICLR}, 2018.

\bibitem{Alemi2017DeepVI}
A.~A. Alemi, I.~S. Fischer, J.~V. Dillon, and K.~Murphy, ``Deep variational
  information bottleneck,'' \emph{ArXiv}, vol. abs/1612.00410, 2017.

\bibitem{Kingma2014AutoEncodingVB}
D.~P. Kingma and M.~Welling, ``Auto-encoding variational bayes,'' \emph{CoRR},
  vol. abs/1312.6114, 2014.

\bibitem{chalk2016relevant}
M.~Chalk, O.~Marre, and G.~Tkacik, ``Relevant sparse codes with variational
  information bottleneck,'' in \emph{Advances in Neural Information Processing
  Systems}, 2016, pp. 1957--1965.

\bibitem{Achille2018InformationDL}
A.~Achille and S.~Soatto, ``Information dropout: Learning optimal
  representations through noisy computation,'' \emph{IEEE Transactions on
  Pattern Analysis and Machine Intelligence}, vol.~40, pp. 2897--2905, 2018.

\bibitem{Amjad2020LearningRF}
R.~A. Amjad and B.~C. Geiger, ``Learning representations for neural
  network-based classification using the information bottleneck principle,''
  \emph{IEEE Transactions on Pattern Analysis and Machine Intelligence},
  vol.~42, pp. 2225--2239, 2020.

\bibitem{Kolchinsky2019NonlinearIB}
A.~Kolchinsky, B.~D. Tracey, and D.~H. Wolpert, ``Nonlinear information
  bottleneck,'' \emph{Entropy}, vol.~21, p. 1181, 2019.

\bibitem{Achille2018EmergenceOI}
A.~Achille and S.~Soatto, ``Emergence of invariance and disentanglement in deep
  representations,'' \emph{2018 Information Theory and Applications Workshop
  (ITA)}, pp. 1--9, 2018.

\bibitem{Yu2020LearningDA}
Y.~Yu, K.~H.~R. Chan, C.~You, C.-B. Song, and Y.~Ma, ``Learning diverse and
  discriminative representations via the principle of maximal coding rate
  reduction,'' \emph{ArXiv}, vol. abs/2006.08558, 2020.

\bibitem{165600}
H.~{Drucker} and Y.~{Le Cun}, ``Improving generalization performance using
  double backpropagation,'' \emph{IEEE Transactions on Neural Networks},
  vol.~3, no.~6, pp. 991--997, 1992.

\bibitem{Xu2011RobustnessAG}
H.~Xu and S.~Mannor, ``Robustness and generalization,'' \emph{Machine
  Learning}, vol.~86, pp. 391--423, 2011.

\bibitem{Rifai2011ContractiveAE}
S.~Rifai, P.~Vincent, X.~Muller, X.~Glorot, and Y.~Bengio, ``Contractive
  auto-encoders: Explicit invariance during feature extraction,'' in
  \emph{Proceedings of the 28th International Conference on International
  Conference on Machine Learning}, ser. ICML'11.\hskip 1em plus 0.5em minus
  0.4em\relax Madison, WI, USA: Omnipress, 2011, p. 833–840.

\bibitem{Gu2015TowardsDN}
S.~Gu and L.~Rigazio, ``Towards deep neural network architectures robust to
  adversarial examples,'' \emph{CoRR}, vol. abs/1412.5068, 2015.

\bibitem{tishby2000information}
N.~Tishby, F.~C. Pereira, and W.~Bialek, ``The information bottleneck method,''
  \emph{arXiv preprint physics/0004057}, 2000.

\bibitem{1405276}
G.~Matz and P.~Duhamel, ``Information geometric formulation and interpretation
  of accelerated blahut-arimoto-type algorithms,'' in \emph{Information Theory
  Workshop}, 2004, pp. 66--70.

\bibitem{Chechik2003InformationBF}
G.~Chechik, A.~Globerson, N.~Tishby, and Y.~Weiss, ``Information bottleneck for
  gaussian variables,'' \emph{Journal of Machine Learning Research}, vol.~6,
  no.~6, pp. 165--188, 2005.

\bibitem{Fischer2020TheCE}
I.~S. Fischer, ``The conditional entropy bottleneck,'' \emph{Entropy}, vol.~22,
  2020.

\bibitem{Wu2019LearnabilityFT}
T.~Wu, I.~Fischer, I.~Chuang, and M.~Tegmark, ``Learnability for the
  information bottleneck,'' \emph{Entropy}, vol.~21, p. 924, 2019.

\bibitem{Rakin2019ParametricNI}
A.~S. Rakin, Z.~He, and D.~Fan, ``Parametric noise injection: Trainable
  randomness to improve deep neural network robustness against adversarial
  attack,'' \emph{2019 IEEE/CVF Conference on Computer Vision and Pattern
  Recognition (CVPR)}, pp. 588--597, 2019.

\bibitem{Oord2018RepresentationLW}
A.~van~den Oord, Y.~Li, and O.~Vinyals, ``Representation learning with
  contrastive predictive coding,'' \emph{ArXiv}, vol. abs/1807.03748, 2018.

\bibitem{Lord2018GeometricKN}
W.~M. Lord, J.~Sun, and E.~M. Bollt, ``Geometric k-nearest neighbor estimation
  of entropy and mutual information,'' \emph{Chaos}, vol. 28 3, p. 033114,
  2018.

\bibitem{Kandasamy2015NonparametricVM}
K.~Kandasamy, A.~Krishnamurthy, B.~Poczos, L.~Wasserman, and j.~m. robins,
  ``Nonparametric von mises estimators for entropies, divergences and mutual
  informations,'' in \emph{Advances in Neural Information Processing Systems},
  C.~Cortes, N.~Lawrence, D.~Lee, M.~Sugiyama, and R.~Garnett, Eds.,
  vol.~28.\hskip 1em plus 0.5em minus 0.4em\relax Curran Associates, Inc.,
  2015, pp. 397--405.

\bibitem{Han2017OptimalRO}
Y.~Han, J.~Jiao, T.~Weissman, and Y.~Wu, ``Optimal rates of entropy estimation
  over lipschitz balls,'' \emph{Ann. Statist.}, vol.~48, no.~6, pp. 3228--3250,
  12 2020.

\bibitem{Wickstrm2019InformationPA}
K.~Wickstr{\o}m, S.~L{\o}kse, M.~Kampffmeyer, S.~Yu, J.~Pr{\'i}ncipe, and
  R.~Jenssen, ``Information plane analysis of deep neural networks via
  matrix-based renyi's entropy and tensor kernels,'' \emph{ArXiv}, vol.
  abs/1909.11396, 2019.

\bibitem{Belghazi2018MutualIN}
M.~I. Belghazi, A.~Baratin, S.~Rajeshwar, S.~Ozair, Y.~Bengio, A.~Courville,
  and D.~Hjelm, ``Mutual information neural estimation,'' in \emph{Proceedings
  of the 35th International Conference on Machine Learning}, ser. Proceedings
  of Machine Learning Research, J.~Dy and A.~Krause, Eds., vol.~80.\hskip 1em
  plus 0.5em minus 0.4em\relax Stockholmsmässan, Stockholm Sweden: PMLR,
  10--15 Jul 2018, pp. 531--540.

\bibitem{Goodfellow2015ExplainingAH}
I.~J. Goodfellow, J.~Shlens, and C.~Szegedy, ``Explaining and harnessing
  adversarial examples,'' \emph{CoRR}, vol. abs/1412.6572, 2015.

\bibitem{MoosaviDezfooli2016DeepFoolAS}
S.-M. Moosavi-Dezfooli, A.~Fawzi, and P.~Frossard, ``Deepfool: A simple and
  accurate method to fool deep neural networks,'' \emph{2016 IEEE Conference on
  Computer Vision and Pattern Recognition (CVPR)}, pp. 2574--2582, 2016.

\bibitem{Nowozin2016fGANTG}
S.~Nowozin, B.~Cseke, and R.~Tomioka, ``F-gan: Training generative neural
  samplers using variational divergence minimization,'' in \emph{Proceedings of
  the 30th International Conference on Neural Information Processing Systems},
  ser. NIPS'16.\hskip 1em plus 0.5em minus 0.4em\relax Red Hook, NY, USA:
  Curran Associates Inc., 2016, p. 271–279.

\bibitem{Donsker1975AsymptoticEO}
M.~Donsker and S.~R.~S. Varadhan, ``Asymptotic evaluation of certain markov
  process expectations for large time,'' \emph{Communications on Pure and
  Applied Mathematics}, vol.~28, no.~1, pp. 1--47, 1975.

\bibitem{Nguyen2017DualDG}
T.~D. Nguyen, T.~Le, H.~Vu, and D.~Phung, ``Dual discriminator generative
  adversarial nets,'' in \emph{NIPS}, 2017.

\bibitem{LeCun1998GradientbasedLA}
Y.~{Lecun}, L.~{Bottou}, Y.~{Bengio}, and P.~{Haffner}, ``Gradient-based
  learning applied to document recognition,'' \emph{Proceedings of the IEEE},
  vol.~86, no.~11, pp. 2278--2324, 1998.

\bibitem{Xiao2017FashionMNISTAN}
H.~Xiao, K.~Rasul, and R.~Vollgraf, ``Fashion-mnist: a novel image dataset for
  benchmarking machine learning algorithms,'' \emph{ArXiv}, vol.
  abs/1708.07747, 2017.

\bibitem{Krizhevsky2009LearningML}
A.~Krizhevsky and G.~Hinton, ``Learning multiple layers of features from tiny
  images,'' \emph{Master's thesis, Department of Computer Science, University
  of Toronto}, 2009.

\bibitem{papernot2018cleverhans}
N.~Papernot, F.~Faghri, N.~Carlini, I.~Goodfellow, R.~Feinman, A.~Kurakin,
  C.~Xie, Y.~Sharma, T.~Brown, A.~Roy, A.~Matyasko, V.~Behzadan,
  K.~Hambardzumyan, Z.~Zhang, Y.-L. Juang, Z.~Li, R.~Sheatsley, A.~Garg,
  J.~Uesato, W.~Gierke, Y.~Dong, D.~Berthelot, P.~Hendricks, J.~Rauber, and
  R.~Long, ``Technical report on the cleverhans v2.1.0 adversarial examples
  library,'' \emph{arXiv preprint arXiv:1610.00768}, 2018.

\end{thebibliography}

%

\end{document}